\documentclass{Interspeech}
\usepackage{multirow}
\usepackage{graphicx}
\usepackage{xcolor}

\interspeechcameraready

\title{Spoken Language Understanding on Unseen Tasks With  In-Context Learning}

\author[affiliation={1}]{Neeraj}{Agrawal}
\author[affiliation={1}]{Sriram}{Ganapathy}

\affiliation{}{Indian Institute of Science, Bangalore}{India}
\email{aneeraj@iisc.ac.in, sriramg@iisc.ac.in}

\usepackage{comment}

\begin{document}

\maketitle

\begin{abstract}
Spoken language understanding (SLU) tasks involve  diverse skills that probe the information extraction, classification and/or generation capabilities of models. In this setting, task-specific training data may not always be available. While traditional task-specific SLU models are unable to cater to such requirements, the speech-text large language models (LLMs) offer a promising alternative with emergent abilities. However, out-of-the-box, our evaluations indicate that the zero/few-shot performance of prominent open-source speech-text LLMs on SLU tasks are not up to the mark. In this paper, we introduce a novel approach to robust task-agnostic fine-tuning  using randomized class labels. With this proposed fine-tuning, we illustrate that the performance of the speech-text LLMs on an unseen task is significantly improved over standard approaches. Critically, the proposed approach avoids the requirement of  task-specific data annotations for enabling new tasks in speech-text LLMs.

\end{abstract}

\section{Introduction}

Although human-computer interactions have become increasingly natural,  Spoken Language Understanding (SLU), the task of uncovering the semantic properties of the speech signal, is one of the key outstanding skills that is still challenging for machines.  
Examples of SLU tasks include, recognizing the intent \cite{lugosch2019speech}, identifying named entities \cite{del2021earnings}, categorizing the sentiments \cite{lu2020speech}, and classifying dialogue acts \cite{ortega2018lexico}. 
While automatic speech recognition (ASR) focuses only on the modality transfer from speech to text, the SLU tasks attempt to directly access the semantic properties of the speech data, thereby enabling  the next-level of multi-modal capabilities in the quest towards artificial general intelligence (AGI). 

The traditional SLU systems often rely on supervised learning with labeled data, which can be expensive and time-consuming to acquire. 
Unlike tasks like speech recognition, the availability of large labeled corpora is a significant challenge in the development of SLU systems. Further, the task is also more complicated than the corresponding text-based semantic reasoning tasks, as speech is multi-layered and nuanced compared to the corresponding text data \cite{arora2022espnet}.  
Most of the successful approaches, until recently, have been based on the cascaded setting, where the ASR system first converts the speech to text, which is followed by a natural language understanding (NLU) module that interprets the semantic properties of the text \cite{peng2023study}. This approach suffers from the cascading of the errors, where the ASR errors as well as the NLU errors directly affect the performance on   SLU tasks. Further, the cascaded setting does not allow the end-to-end optimization of the models for the SLU objective. 
Recently, end-to-end SLU systems \cite{saxon2021end} have been attempted to directly infer the SLU  labels from the speech data. 

\begin{figure*}[t!]
  \centering
    \includegraphics[width=\textwidth]{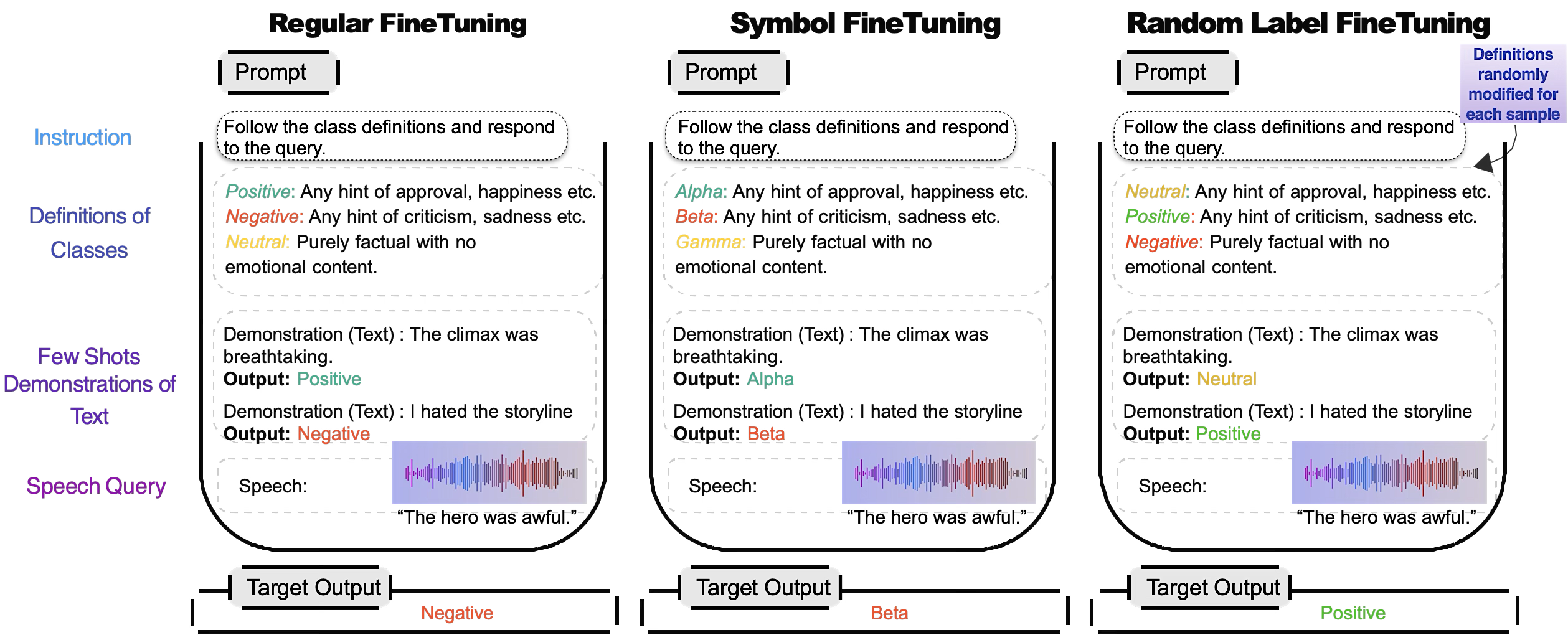}
  \caption{Examples of different strategies for fine-tuning including regular fine-tuning, symbol based fine-tuning and random label fine-tuning. }
  \label{fig:main_arch}
  \vspace{-0.1in} 
\end{figure*} 
On the other hand, development of  large language models (LLMs) have substantially improved the semantic reasoning and NLU capabilities of text-based models.  
 Although text-based LLMs show emergent abilities in zero-shot settings, in the scheme of multi-modal speech-text LLMs, we illustrate that  they exhibit limited zero-shot SLU capabilities.  In such a scenario, this paper explores the direction of in-context-learning (ICL) aided with limited (and task mis-matched) fine-tuning to enable novel SLU tasks with speech-text LLMs. 
 The approach of ICL  \cite{dong2024survey,zhou2023mystery, luo2024context} allows models to learn new tasks with the aid of only a  few examples \cite{luo2024context,unknown}. 

The ICL capabilities have largely been shown for text-based tasks \cite{fei2023mitigating,workrethinking}, while multi-modal ICL (M-ICL) has proved to be more challenging \cite{doveh2024towards, alayrac2022flamingo,chen2024visual}. These works show that M-ICL heavily relies on text-driven mechanisms, with minimal influence from the other modalities. It is also observed many open-source smaller models have limited M-ICL capabilities compared to large black-models like GPT-4o and Gemini \cite{jiang2024many}. 




With the exception of a small number of studies \cite{chang2024exploring},  M-ICL has not been explored in the past for SLU tasks using speech-text LLMs. However, we argue that the ICL capability is particularly beneficial for the SLU setting, as it often involves diverse and complex tasks  with limited supervised task-specific data. 
 Additionally, very few past efforts have focused on the cross task capabilities \cite{chatterjee2024language, cheng-etal-2023-uprise} and the ability to generalize to unseen tasks \cite{min2021metaicl,wang2022super}. 
 In this paper, we develop ICL strategies for multi-modal (audio-text) unseen tasks with symbolic based fine-tuning and randomized label fine-tuning. 
 We illustrate that the random label based fine-tuning enables the few-shot capabilities of the speech-text LLM on a completely unseen task. 


\section{Related Work}
\textbf{In context learning (ICL)}: The field of in-context learning (ICL) has witnessed significant progress, particularly with the demonstration that large language models (LLMs) can perform few-shot learning without explicit fine-tuning, simply by conditioning on context \cite{DBLP}. This paradigm has been extended to multimodal settings, with models like Flamingo demonstrating effective ICL by conditioning on both textual and visual inputs \cite{alayrac2022flamingo}.  In contrast to these works which focus on text or vision, our research explores ICL specifically for spoken language understanding, leveraging both text and audio modalities.


\noindent \textbf{Zero shot and few-shot evaluation of SLU}: Zero-shot probing of SLU tasks has gained prominence recently, with methods like cross-modal selective self-training \cite{he2023zero}.  
Few-shot ICL has also demonstrated potential, with studies comparing ICL to supervised fine-tuning and instruction tuning \cite{razumovskaia2024analyzing} and investigating the generalization capabilities of few-shot fine-tuning.  Unlike these studies, our work explores a novel approach to improve generalization across tasks and modalities.

\noindent \textbf{Emergent Abilities For Novel Tasks}: Recent work has explored ICL for speech classification using textless speech language models \cite{chang2024exploring}, improving ASR in low-resource languages with Meta-ICL \cite{hsu2024meta}, and evaluating Whisper's ability to perform ICL directly with speech input \cite{wang2024can, he2023can}. Other research has focused on using LLMs for SLU tasks with textual input \cite{he2023can} and data-efficient instruction tuning of speech models  \cite{pan2023cosmic}. Cross-task ICL has also been investigated, with approaches involving training small LLMs to retrieve prompts \cite{cheng-etal-2023-uprise} and evaluating zero-shot cross-task performance with multiple LLMs \cite{chatterjee2024language}.  In contrast, our work introduces a novel fine-tuning approach to  address the combined challenges of cross-task and cross-modal generalization in speech-text LLMs.
Our work builds upon previous research on symbol tuning for ICL \cite{wei2023symbol} and instruction tuning for multi-modal tasks \cite{li2023instruction}, extending these concepts to the domain of multi-modal task-agnostic ICL.

\renewcommand{\arraystretch}{1.3}
\begin{table*}[h]
    \centering
    \begin{tabular}{|c|c|l|c|c|c|c|c|c|c|}
        \toprule
         \multirow{2}{*}{\begin{tabular}[c]{@{}c@{}}\textbf{Eval.} \\ \textbf{Task} ($\boldsymbol{T_E}$)\end{tabular}} &\multirow{2}{*}{\begin{tabular}[c]{@{}c@{}}\textbf{FT} \\ \textbf{Task} ($\boldsymbol {T_{FT}}$)\end{tabular}} & \multirow{2}{*}{\begin{tabular}[c]{@{}c@{}} ~~~~\textbf{FT} \\~~~~~~~~\textbf{Strategy}\end{tabular}}   & \multicolumn{6}{c|}{\textbf{\# of Demonstrations in Eval.}} \\
         \cline{4-9}
         &  & &  \textbf{0 }& \textbf{1} & \textbf{2} & \textbf{3} &\textbf{ 4} &\textbf{ 5}  \\
        \midrule
        \multirow{8}{*}{HVB} &\multicolumn{2}{c|}{\multirow{1}{*}{~~No FT}}  & 1.3 & 31.6 & 22.2 & 14.9 & 14.6 & 12.5  \\
        
        \cline{2-9}
        & \multirow{3}{*}{HVB} & \multirow{1}{*}{\begin{tabular}[c]{@{}c@{}}Regular  FT\end{tabular}}  & 38.0 & 50.0 & 51.2 & 52.2 & \underline {52.6} & 52.5    \\
        
        &  & \multirow{1}{*}{\begin{tabular}[c]{@{}c@{}}Symbol FT\end{tabular}}  & 20.9 &34.6  & 38.1 & 40.0 &43.9  &45.6    \\
        &  & \multirow{1}{*}{\begin{tabular}[c]{@{}c@{}} Random Label FT\end{tabular}}  &14.1&16.7 &23.6  & 31.9& 38.9& 45.8 \\
       
        \cline{2-9}
        &  \multirow{3}{*}{Voxceleb} &\multirow{1}{*}{\begin{tabular}[c]{@{}c@{}}Regular  FT\end{tabular}}  & 0.8 & 24.2  & 22.7 & 23.5  & 21.7 & 19.3  \\
     
        &  & \multirow{1}{*}{\begin{tabular}[c]{@{}c@{}}Symbol FT\end{tabular}}  & 0.8 &25.2 &  21.7 &22.8 & 22.4 & 22.6  \\
        
        &  & \multirow{1}{*}{\begin{tabular}[c]{@{}c@{}} Random Label FT\end{tabular}}  &\textbf{3.8}  & \textbf{41.1}  &\textbf{39.8}  &\textbf{42.2}  & \textbf{45.2} & \textbf{47.0 }
       \\ \midrule \midrule

        \multirow{8}{*}{Voxceleb} &\multicolumn{2}{c|}{\multirow{1}{*}{~~No FT}}  & 36.2 & 16.9 & 20.7 & 24.1 & 25.0 & 28.3  \\
       
        \cline{2-9}

          &\multirow{3}{*}{Voxceleb} & \multirow{1}{*}{\begin{tabular}[c]{@{}c@{}}Regular FT\end{tabular}} & 61.1 & 64.3 & 63.3 &  63.6&  64.3& \underline{64.7}      \\
     
        &  & \multirow{1}{*}{\begin{tabular}[c]{@{}c@{}} Symbol FT\end{tabular}}  &52.0  & 63.4  &63.0  & 63.4  & 64.0 & 63.2 \\
      
        & &   \multirow{1}{*}{\begin{tabular}[c]{@{}c@{}} Random Label FT\end{tabular}}   & 46.4  & 56.7  & 54.2 & 53.8 & 53.7 & 54.3 \\
       
         \cline{2-9}
         & \multirow{3}{*}{HVB}  &\multirow{1}{*}{\begin{tabular}[c]{@{}c@{}}Regular FT\end{tabular}} & 34.5& 31.3 & 35.8 & 39.1 &  41.0&42.3   \\

         & & \multirow{1}{*}{\begin{tabular}[c]{@{}c@{}}Symbol  FT\end{tabular}} & \textbf{42.8 }  & 35.8 & \textbf{41.3}  & 43.7  & \textbf{46.1}   & 45.1    \\
        
        & &  \multirow{1}{*}{\begin{tabular}[c]{@{}c@{}}Random  Label  FT\end{tabular}} & 36.2 & \textbf{36.0}  & 40.4   & \textbf{43.9}  & 44.5   & \textbf{45.8}    \\

        \midrule \midrule
       \multirow{4}{*}{Voxpopuli} &\multicolumn{2}{c|}{\multirow{1}{*}{~~No FT}}   & 0.4 & 4.3 &2.4 & 1.4 & 1.0& 1.1   \\
        
        \cline{2-9}
        & \multirow{3}{*}{Voxceleb}  & \multirow{1}{*}{\begin{tabular}[c]{@{}c@{}} Regular FT\end{tabular}}  & 7.9 & 11.1  & 11.9 & 12.5 & 13.0 & 11.4\\


        & & \multirow{1}{*}{\begin{tabular}[c]{@{}c@{}} Symbol FT\end{tabular}}  &4.1  & 17.4  &19.3  & 18.1  &17.4 & 17.3\\

        & &   \multirow{1}{*}{\begin{tabular}[c]{@{}c@{}} Random Label FT\end{tabular}}  & \textbf{ 8.3} & \textbf{17.9} & \textbf{19.3} & \textbf{22.1} &\textbf{ 22.3 }& \textbf{ 21.5  } \\

        \bottomrule
    \end{tabular}
    \vspace{0.1in}
    \caption{Table shows different fine-tuning (FT) strategies for cross task learning and cross modal ICL. In all cases, the query is speech (S), while the demonstrations are in text (T). The focus of the work is on generalization towards unseen/mis-matched tasks ($T_E \neq T_{FT}$). Hence, the best performance on a given SLUE task in an mis-match setting is given in \textbf{bold}. The best performance among the few-shot cases in the matched setting ($T_E = T_{FT}$) is highlighted in \underline{underline}.  }
    \label{tab:my_label}
    \vspace{-0.25in}
\end{table*}

\section{Fine-Tuning Strategies}
We explore different strategies (depicted in Figure~\ref{fig:main_arch}) for fine-tuning speech-text LLM models.
\subsection{Overall Setting}
In all the settings, the instructions and class-definitions are given in textual form while the final query is provided in speech form.  
There are two distinct tasks involved - i) the fine-tuning task ($T_{FT}$), where the data with the annotations are used to update the speech-text LLM model parameter weights  and ii) the evaluation task ($T_E$), where the performance is measured. 
In a match setting, both the fine-tuning and evaluation tasks are identical, while in a mis-match setting, the evaluation task is different from the fine-tuning task and is unseen. 
 For the performance evaluation, we explore zero-shot as well as few-shot prompting. The few-shot prompting presents a small number of examples with correct responses to guide the model to perform the task at hand. 
We refer to the few-shot examples   as ``Demonstrations'' ($D$) to the model. 
We experiment with demonstrations in textual form  (using transcriptions of the audio), denoted as $D(T)$. In the fine-tuning stage, a random number of demonstrations in text are also appended along with the speech-prompts.   
We use the SALMONN speech-text LLM~\cite{tang2024salmonn} in all our experiments and fine-tune the speech encoder, Q-former parameters as well as the LoRA parameters in the LLaMa LLM model. 
Also, due to the context size (with demonstrations) and model size (SALMONN-13B) limitations, all the fine-tuning experiments were performed with a batch size of $1$. 

\subsection{Regular Fine-tuning}
In the regular fine-tuning, we follow the instructions using the definition of the classes given in the fine-tuning task directly. This is shown in the left panel of the Figure~\ref{fig:main_arch}. The model learns the mapping between the speech queries (along with the text demonstrations) and the class-labels. 
During the evaluation, with a mis-match task, the model is required to generalize to a novel task with a new-set of target classes.  
The regular fine-tuning is the most common mode of supervised fine-tuning utilized in LLMs.

\subsection{Symbol Fine-tuning}
The class names (like ``positive'' or ``neutral'') have semantic associations with English words. Thus, the regular fine-tuning inherently combines the representations of the class names with the model's internal representations  that have been learned during pre-training. 
To avoid this composition of class-names and their semantic counter-parts, we define the class names symbolically ($alpha$, $beta$ etc). This is inspired by symbol-tuning   work by Wei et. al \cite{wei2023symbol}. In the matched evaluation setting, the class definitions are reverted back to the original definitions. While this creates a difference in matched setting between the class-definitions, we hypothesize that symbolic definitions allow an improved cross-task transfer of ICL capabilities. This setting is shown in the middle panel of Figure~\ref{fig:main_arch}. 

\subsection{Random Label Fine-tuning}
We propose a novel framework for fine-tuning to improve task agnostic generalization of speech-text LLM capabilities to novel unseen tasks in the mis-matched setting. 
The goal of this strategy is to enable the model to follow the instructions carefully for each sample and to facilitate the abstraction of the task requirements beyond the specific fine-tuning task being solved. 
By striving to achieve this goal, the speech-text LLM model is more flexible to follow specific instructions at evaluation which may differ from those provided during the fine-tuning stage.  
The random label fine-tuning also forces the model to ignore any prior semantic definitions of the class-labels. 

In order to achieve this goal, we propose to randomly permute the definition of classes at every mini-batch of fine-tuning. For example, in one mini-batch, the definition of ``neutral'' class is \textit{any hint of approval or happiness}, while in another mini-batch the definition is of ``neutral'' class is modified to \textit{any hint of criticism or sadness}. 
Note that, the class-definition for a given example is uniform across the text demonstration (few shot examples $D(T)$) and the speech query. During evaluation, the class definitions pertaining to the task are directly used. This setting is shown in the right panel of Figure~\ref{fig:main_arch}.

\section{Data}
We have evaluated our approach on three tasks,  SLUE VoxCeleb \cite{shon2022slue}, the SLUE-HVB \cite{shon-etal-2023-slue} and SLUE-VoxPopuli \cite{shon2022slue}. \\
\noindent  \textbf{SLUE-Voxceleb:} SLUE-Voxceleb is based on audio extracted from YouTube videos, collected as part of the VoxCeleb corpus \cite{shon2022slue}. In this dataset, each spoken utterance is labeled with one of three sentiment classes: positive, negative, and neutral. To assess the sentiment analysis   performance, we calculate macro-averaged F1 score. Hence, this task is a $3$-class classification setting. For this task, the model is fine-tuned on $5,777$ samples and evaluated on $3,553$ samples. This dataset contains $1454$ validation examples.

\noindent \textbf{SLUE-HVB:} HarperValley Bank corpus consists of scripted dialogues between bank employees and customers \cite{shon-etal-2023-slue}. The dialog act labels in SLUE-HVB include $18$ actions (fine-grained labeling scheme). Each recording can also have multiple actions.  We evaluate the speech-text LLMs on their ability to classify fine-grained labels  using macro-averaged F1 score. 
Hence, this setting is a multi-label $18$-class classification setting and thus, inherently more complex compared to the SLUE-Voxceleb task. There are $11,344$ samples  for fine-tuning and   $6,121$ samples for evaluation. HVB dataset have $1690$ validation samples.

\noindent \textbf{SLUE-VoxPopuli:} The Voxpopuli dataset for SLUE \cite{shon2022slue} relates to the task of Named Entity Recognition (NER). NER is the process of identifying named entities in a sentence and assigning them with appropriate labels (types). Named entities are phrases—often, composed of proper nouns—that represent specific entities such as individuals, locations, organizations, numerical values, and more. Model is evaluated on $7$ combined NER entity tags. Similar to HVB this is also multi-label classification task. We use the micro-averaged F1 score to measure the performance on the SLUE-VoxPopuli dataset. The evaluation data consists of $1,843$ speech recordings. VoxPopuli contains $1753$ samples in validation split. We use the SLUE-VoxPopuli only in the evaluation and do not perform any fine-tuning on this dataset.\\

\section{Experimental Setup}
\subsection{Prompt Structure and fine-tuning Procedure}
We employed a consistent prompt structure across all fine-tuning strategies, consisting of:
\begin{itemize}
    \item \textbf{Task Instruction:} A clear instruction defining the task, such as "\texttt{You are a dialogue analysis expert. Based on the statement below, identify all applicable dialogue actions from the given categories.}"

    \item \textbf{Class Definitions:} Precise definitions of each categorical class pertinent to the task.
    \item \textbf{General Guidelines:} Instructions regarding output format, the number of outputs (single or multiple), and other task-specific guidelines.
    \item \textbf{Few-shot demonstrations:} A variable number of few-shot demonstrations presented in text format with text labels.
    \item \textbf{Input Speech Query:} The actual speech input which the model uses to predict the output labels.
    \item \textbf{True Label:} The corresponding ground truth label for the input query (used in fine-tuning).
\end{itemize}

All speech queries and speech demonstrations are enclosed within the special tokens \verb|<Speech>| and \verb|</Speech>|. During the fine-tuning stage, the number of few-shot demonstrations are randomly selected between $0$ and $5$. 
The demonstrations for few-shot prompting are obtained using text-embedding semantic similarity score between query under consideration (textual form of speech query is available in the dataset)  with the textual form of training samples  \cite{luo2024bgelandmarkembeddingchunkingfree}. The ASR transcripts of the speech query and training examples would also allow the  few-shot example creation. 

For the VoxCeleb SLU random label fine-tuning, we created six training combinations by swapping each label definition, while for the HVB dataset, we generated ten random label definitions. During inference, the same prompt structure is used, but with a different set of few-shot examples choosen from train and validation dataset based on the embedding similarity score.

\subsection{Fine-tuning Details}
We employed LoRA with a rank ($r$) of $8$, alpha ($\alpha$) of $32$, and a dropout rate of $0.1$ \cite{Hu2021LoRALA}. The cross-entropy loss was calculated between the predicted output tokens for labels and the target labels. We optimized the model using Adam with an initial learning rate of $1e-5$, warm-up steps of $100$, and a cosine annealing learning rate scheduler. Gradient accumulation and gradient clipping were implemented to expedite training and prevent gradient explosions. Each model was trained for $15$ epochs.

\section{Results and Discussion}
Table \ref{tab:my_label} presents the performance of the SALMONN architecture under various experimental conditions, focusing on the impact of fine-tuning strategies   on cross-task and cross-modal in-context learning (ICL). We compare the performance on various combinations of fine-tuning ($T_{FT}$) and evaluation ($T_E$) tasks. The SLUE-VoxPopuli setup is only used in the evaluation, where the model was fine-tuned in SLUE-Voxceleb task. 

The following are the key-takeaways from the experimental analysis.
\begin{itemize}
    \item The performance in the no-fine-tuning (No FT) for all the tasks is significantly inferior to those achieved with match/mis-match fine-tuning. The No FT results form the lower bound for the performance on the task. This trend motivates the need for  SLU fine-tuning even if the fine-tuning task is mis-matched to the evaluation. 
    \item In all the experiments,  both matched and mismatched, the few-shot performance improves over the zero-shot case. This improvement  emphasizes the value of ICL for SLU tasks. 
    \item While stating the obvious, the matched FT provides the best performance in SLUE-HVB and SLUE-Voxceleb evaluations with the regular FT strategy. These result form the upper-bound of the task performance. The few-shot performance is also seen to improve the performance over the zero-shot setting. 
    \item In mis-matched settings, three cases are considered - i) $T_E$=SLUE-HVB, with $T_{FT}$=SLUE-Voxceleb,  ii) $T_E$=SLUE-Voxceleb, with $T_{FT}$=SLUE-HVB,  and  iii) $T_E$=SLUE-Voxpopuli, with $T_{FT}$=SLUE-Voxceleb. \\ \textbf{The random-label FT provides the best mis-matched performance on most of the zero-shot and few-shot settings. }
    
\end{itemize}
 In case of mis-matched evaluation with $T_E$=SLUE-HVB, the random-label FT achieves a performance that is $21.6$\% relatively lower than the upper-bound (matched setting with regular FT), while in the case of  mis-matched evaluation with $T_E$=SLUE-Voxceleb, the random-label FT achieves a performance that is $54.5$\% relatively lower than the upper-bound. In all cases, the random-label FT significantly improves over the No-FT as well as the regular-FT approaches. On the average, over all the few-shot cases, the random label fine-tuning provides a relative improvement of $10.1$\% on the SLUE-Voxceleb mis-match evaluation over the regular fine-tuning, while providing a relative improvement of $95.3$\% over the regular fine-tuning on the SLUE-HVB mis-match evaluations. 
    On the SLUE-Voxpopuli evaluations, the random fine-tuning improves by $64.3$\% over the regular fine-tuning.  To the best of our knowledge, our proposal of random label fine-tuning, to enable task-agnostic learning, is explored for the first time in LLMs.  The experimental results show that, a randomized label based fine-tuning strategy on a mis-matched FT task, enables the unlocking of emergent abilities of the models to unseen SLU tasks.

\section{Conclusion}
In this paper, we considered the emergent capabilities of speech-text LLMs on SLU tasks that are unseen during training and fine-tuning. For such a setting, we explored multiple strategies with a SALMONN based speech-text model. In order to avoid the dependence on the training-labels used in a regular fine-tuning setup, we proposed a symbol based fine-tuning and a randomized label based fine-tuning strategy. 
Extensive evaluation experiments were performed on $3$ SLU tasks in matched and mis-matched settings. We performed both zero-shot as well as few-shot (textual few-shot examples in context) evaluations.
In these experiments, we illustrate that the randomized label fine-tuning provides the best performance, while significantly improving over the regular fine-tuning or no-fine-tuning approaches. 
The future efforts in our work would attempt to develop the theoretical understanding of randomized label-based fine-tuning in the context of LLMs and to extend the empirical study to a wider range of data and tasks. 


\bibliographystyle{IEEEtran}
\bibliography{mybib}

\end{document}